\documentclass[letterpaper, 10 pt, journal, twoside]{IEEEtran}
\usepackage{amsmath,amsfonts}
\usepackage{xcolor}
\usepackage{algorithm}
\usepackage{algpseudocode}
\usepackage{array}
\usepackage{textcomp}
\usepackage{stfloats}
\usepackage{url}
\usepackage{verbatim}
\usepackage{cite}
\usepackage{eqexpl}
\usepackage[normalem]{ulem}
\usepackage{subcaption}
\usepackage{graphicx}
\usepackage[hidelinks]{hyperref}

\newcommand\planeParameters{\mathbf{P}}
\newcommand\dplaneParameters{\mathbf{\Delta P}}

\begin{document}

\title{Humanoid Robot Whole-body Geometric Calibration with Embedded Sensors and a Single Plane}

\author{Thanh D. V. Nguyen$^{1,2,*}$, Vincent Bonnet$^{1}$, Pierre Fernbach$^{2}$, David Daney$^{3}$, Florent Lamiraux$^{1}$ % <-this % stops a space
\thanks{${^1}$ LAAS-CNRS, Université Paul Sabatier, CNRS, Toulouse, France}
\thanks{$^{2}$ TOWARD S.A.S, Toulouse, France}
\thanks{$^{3}$ INRIA Bordeaux, Bordeaux, France}
\thanks{$^{*}$ corresponding author: {\tt\small dvtnguyen@laas.fr}}
\thanks{This work was partially supported by PAL Robotics}% <-this % stops a space
}
% The paper headers
% Paper headers
\markboth{IEEE Transaction on Automation Science and Engineering. Preprint Version. Accepted Month, Year}
{FirstAuthorSurname \MakeLowercase{\textit{et al.}}: ShortTitle} 
% Use only for final RAL version
% \IEEEpubid{0000--0000/00\$00.00~\copyright~2021 IEEE}
% \IEEEpubidadjcol
% Remember, if you use this you must call \IEEEpubidadjcol in the second
% column for its text to clear the IEEEpubid mark.
% \IEEEpubid{0000--0000/00\$00.00 \copyright\ 2013 IEEE}

\maketitle

\begin{abstract}

% Whole-body geometric calibration of humanoid robots using classical robotics methods is very time consuming and experimentally cumbersome. However, even if too often not considered by the community in humanoid robotics, it is required for accurate control and simulation.
% Therefore, we propose a new practical method based on a single plane, embedded force sensors and an admittance controller to calibrate, without manual intervention, the whole-body kinematics of humanoids. As humanoids have numerous joints, it is also crucial to generate and determine a minimal set of optimal calibration poses.
Whole-body geometric calibration of humanoid robots using classical robot calibration methods is a time-consuming and experimentally burdensome task. However, despite its significance for accurate control and simulation, it is often overlooked in the humanoid robotics community. To address this issue, we propose a novel practical method that utilizes a single plane, embedded force sensors, and an admittance controller to calibrate the whole-body kinematics of humanoids without requiring manual intervention. Given the complexity of humanoid robots, it is crucial to generate and determine a minimal set of optimal calibration postures. To do so, we propose a new algorithm called IROC (Information Ranking algorithm for selecting Optimal Calibration postures). IROC requires a pool of feasible candidate postures to build a normalized weighted information matrix for each posture. Then, contrary to other algorithms from the literature, IROC will determine the minimal number of optimal postures that are to be played onto a robot for its calibration. Both IROC and the single-plane calibration method were experimentally validated on a TALOS humanoid robot. The total whole-body kinematics chain was calibrated using solely 31 optimal postures with 3-point contacts on a table by the robot gripper. In a cross-validation experiment, the average RMS error was reduced by a factor of 2.3 compared to the manufacturer’s model.

\end{abstract}

% \begin{IEEEkeywords}
% Calibration and Identification, Humanoid Robot Systems, Plane-constrained Calibration
% \end{IEEEkeywords}

%Note: \textcolor{red}{red text}: removed text \textcolor{blue}{blue text}: replaced text.

\section{Introduction}

% * Interest in humanoid robots lead to the need of  calibration for humanoids

Humanoid robots can now be seen outside of the laboratories with companies that are already offering commercial versions for various applications~\cite{digit_robot}.
During the development phase or when undergoing modifications, such as to meet new experimental requirements, humanoids are often dismantled, repaired, and reassembled. However, this process leads to an inaccurate kinematic model of the robot. Geometric errors refer to the slight 3D position and 3D orientation variations of the nominal values between two consecutive frames. 
% In humanoids, as in any robot, the end-effector, i.e. the hand, pose is the most important, but it can be effectively achieved only if the pose of the floating base and thus of the feet are also well-estimated. The floating base pose  usually relies on the forward kinematic model (FKM) and IMU data that are, after a static transformation from the IMU frame to the floating base one, fused together in an adaptive filter~\cite{Rotella14}. If these models are incorrect, so will be the pose estimation of each link. Thus the calibration of the hand pose must include the whole kinematics chain starting from the feet. 
In humanoids, as well as in any robot, achieving an accurate end-effector pose, specifically the hand pose, is crucial. However, this can be effectively accomplished only if the pose of the floating base, and thus of the feet, are also accurately estimated. Typically, the floating base pose is determined through a combination of the forward kinematic model (FKM) and IMU data. These inputs are fused together in an adaptive filter after a static transformation from the IMU frame to the floating base frame~\cite{Rotella14}. If these models are inaccurate, the estimation of each link pose will also be affected. Hence, when calibrating the hand pose, it is essential to consider the entire kinematic chain, starting from the feet.

% To calibrate a robot, classical methods proposed to use very accurate Laser trackers~\cite{JOUBAIR2013_compare_index} or similar external measurement tools that can measure only the 3D position or pose of a single point for a given posture.  A humanoid is intrinsically unstable, thus for each calibration posture the pose of the base of the robot must be re-measured. As a humanoid is composed of thirty or more joints and because calibration requires much more calibration postures than joints, the calibration of the complete kinematic model using standard Laser tracker would be very cumbersome.  
In robot calibration, classical methods have relied on highly precise laser trackers or similar external measurement tools capable of measuring the 3D position or pose of a single point in a specific posture~\cite{JOUBAIR2013_compare_index}. However, when calibrating a humanoid robot, which inherently lacks stability, it becomes necessary to re-measure the base pose for each calibration posture. Considering that a humanoid typically consists of thirty or more joints, the sheer number of calibration postures required for a complete kinematic model calibration would make the process exceedingly cumbersome if relying solely on standard laser trackers.
In this context, there is a need to develop new practical methods that allow for the reduction of the experimental burden and time required to identify the geometric parameters of humanoid robots. 
\section{Paper contributions}
The main contributions of this paper are as follows:

\begin{itemize}
    \item A complete framework for practical whole-body humanoid geometric calibration to identify a full set of geometric parameters without external sensors.
    \item A new algorithm, called IROC (Information Ranking algorithm for selecting Optimal Calibration postures), to generate an optimal set of calibration postures, simplifying the process of solving the combinatorial optimization problem in robot calibration experiment design.
    \item An experimental validation of the whole-body geometric calibration with a TALOS humanoid robot~\cite{TALOS2017}, based on solely embedded sensors and a single plane.
\end{itemize}

The paper is organized as follows. After reminding previous works in Section~\ref{sec:related-work}, Section~\ref{sec:methodology} describes the TALOS calibration model using data gathered from a single plane and the IROC algorithm. Section~\ref{sec:results} presents the IROC algorithm output and the experimental validation obtained with the TALOS humanoid robot. Finally, the paper is concluded  by a discussion about results and future works.

\section{Related work}\label{sec:related-work}

%In classical robot modeling, Denavith-Hartenberg convention has been proposed to represent with  four identifiable parameters  any transformation from one joint to its neighbor joint. In more recent years, the relative actual 3D position and 3D orientation are more preferred in standard modeling formats such as URDF. As a result, either one of the two conventions can be chosen as parameters to formulate a geometric calibration problem.

Industrial robot geometric calibration has been the subject of numerous studies since the early 2000s. Various authors have extensively investigated different sensor modalities, modeling techniques, and criteria for assessing and improving the calibration quality of 6 Degrees-of-Freedom (DoF) serial manipulators. However, the methods developed for industrial robots need to be adapted for humanoid robots. This is primarily due to the complex kinematic structures of humanoid robots, making the use of external sensors such as laser trackers or optical coordinate measuring machines impractical~\cite{bonnet2023}. Additionally, algorithms designed for generating Optimal Calibration Postures (OCP) are not well-suited for the numerous DoF present in humanoid robots, often resulting in slow computation times and the risk of converging to local minima~\cite{Daney2005}. Remarkably, despite these theoretical and practical challenges, the calibration of humanoid robots has not received significant attention in the field.

\subsection{Humanoid robot calibration}

    The floating base pose of humanoids should be measured and considered in the calibration model.  Alas, this is often neglected in most  studies that focus solely on upper or lower segments using simplified calibration models and embedded sensors .

    Birbach et al.~\cite{Birbach2015}, for instance, considered only joint angle offsets and joint elasticities of upper segments using data collected from two cameras attached on the head and tracking markers attached on the end-effectors. Although the authors combined camera data with measurements from an IMU and a Kinect sensor, the accuracy of the measurements and the calibration process was restricted and complex to understand since the calibration results were presented in the form of camera poses with respect to the head frame.
    Others~\cite{Roncone2014, Stepanova2019} proposed to calibrate the full set of kinematic parameters of the upper segments of an iCub robot using data from a camera mounted on iCub's head and from self-contact location obtained using a new tactile sensing skin. Although this work offered to calibrate the full set of kinematic parameters, the study~\cite{Stepanova2019} was done only in simulation, and as mentioned by the authors, the accuracy of tactile skin is still to be determined~\cite{Roncone2014}.  
  
Recently,    Stepanova et al. \cite{Stepanova2022} compared different self-contained calibration approaches such as: self-contact, self-observation with embedded camera, planar constraints and external calibration using high-accuracy laser tracker system on a dual-arms platform. It was observed by the results that while self-contact approach would yield to the best results, combining multiple approaches together into one single cost function would even lower the errors and improve the observability. The study showed that self-contained approaches with on-board sensors are viable alternatives to external measurement systems for robot calibration.
    
    For lower segment calibration, Yamane et al.~\cite{Yamane2011} presented a method to calibrate joint angle offsets with data obtained from an IMU mounted to the pelvis link, and with information from geometric constraints that kept two feet of the robot fixed on a floor. The experimentation performing calibration postures involved manually moving the robot legs, which made this approach quite cumbersome. In another approach~\cite{Khusainov2017}, the leg calibration process was done by having an industrial manipulator moving the foot, while the base of the humanoid was fixed with respect to the industrial manipulator. Although this approach is intriguing, it can be expensive and challenging to automate for all kinematics chains, and its practical validation was limited to a small humanoid robot.  
    
    To the best of our knowledge, whole-body calibration was investigated only in two studies. Tanguy et al.~\cite{Tanguy2018} proposed a SLAM-based algorithm to extend the eye-hand calibration to an online  whole-body calibration of geometric parameters using only embedded sensors. However, no quantitative evaluation was provided to demonstrate the efficiency of this method. The second approach, recently proposed by our group~\cite{bonnet2023}, proposed a whole-body elasto-geometric calibration for a TALOS robot using a stereophotogrammetric system. In this setup, several markers were mounted on various locations of the robot and were tracked by a system of 20 cameras. Using the recorded location of these markers, the authors identified a full-set of kinematic parameters and joint elasticities of the robot legs. However, the performance of the calibration work relied significantly on the accuracy of the tracking and measuring of the stereophotogrammetric system. To ensure the measuring accuracy, a rather lengthy additional calibration for the cameras are required before every experiment. In this context, it appears more suitable to use embedded sensors only to identify the full set of geometric parameters. 
    % revise wording and grammars: 17/07/23 - Thanh
    
    %In addition to that, Maier et al.~\cite{Maier2015} also proposed to extend the eye-hand calibration to identify joint angle offsets of a whole-body NAO robot. In this work setup, a monocular camera on the head of NAO observed markers attached on the end-effectors such as wrists and feet. However, as mentioned earlier, eye-hand calibration might not be ideal for high accuracy calibration, especially for humanoids.

\subsection{Plane-constrained calibration}

    % Besides using embedded sensors and external optical measurement systems, the literature proposed several geometry-constrained calibration methods~\cite{khalil:hal-00362605, 606774, 770073, Joubair2015NonkinematicCO, YANG2020265}. Given the geometry of the constraints on which the desired motion of the end-effector is restricted, it is possible to obtain one or more positional and/or orientational measurements. 
    % % This approach is interesting as there is no need for external measuring system and it can be automatized, thus reducing the overall calibration time and burden. 
    % Among geometry-constrained calibration methods, plane-constrained calibration is one of the most popular. Most approaches in planar constraint calibration method usually use a trigger probe which makes single-point contacts with one or multiple planes~\cite{khalil:hal-00362605,606774,770073}.
    % Zhuang et al.~\cite{770073} pointed out that when using single-point planar contacts, at least 3 orthogonal planes are required to obtain the same amount of information as when using optical measuring systems.
   In addition to utilizing embedded sensors and external optical measurement systems, various geometry-constrained calibration methods have been proposed in the literature~\cite{khalil:hal-00362605, 606774, 770073, Joubair2015NonkinematicCO}. These methods leverage the geometry of constraints to obtain positional and/or orientational measurements.
    Among these methods, plane-constrained calibration stands out as one of the most widely used. In this approach, a trigger probe is often employed to make single-point contacts with one or multiple planes~\cite{khalil:hal-00362605,606774,770073}.
    Zhuang et al.~\cite{770073} noted that when employing single-point planar contacts, at least three orthogonal planes are necessary to acquire the same level of information as that obtained from optical measuring systems.
    
    To our best knowledge, plane-constrained calibration has only been explored once in the context of humanoids, as presented in a preliminary study involving a simple joint offset calibration model~\cite{Yamane2011}. In this study, manual re-positioning of the feet on a plane was conducted for each new posture, with measurements based on the constant height between two feet vertices. Consequently, an automated process for plane-constrained calibration in humanoid robots, capable of achieving accuracy equivalent to that of external optical measurement systems without relying on external sensors, would be highly valuable for the community.

\subsection{Optimal calibrating posture generation}\label{subsection:select_ocp}
    \subsubsection{OCP selection methods}
    Irrespective of the measurement system used, OCP are essential for enhancing accuracy while reducing experimental time and workload~\cite{bonnet2023}. This can be achieved through a non-linear optimization problem optimizing an excitation index while ensuring mechanical and collision constraints~\cite{Daney2005}. However, determining an optimal and feasible set of calibrating postures for humanoid robots remains an ongoing challenge~\cite{bonnet2023}. A practical approach involves selecting optimal postures from a large pool of feasible postures obtainable using simple inverse kinematics~\cite{Daney2005}. The selection process is often perceived as solving a combinatorial optimization problem.
    
    A common approach to solve this problem is to implement a greedy search in the feasible pool of candidates such as the DETMAX algorithm~\cite{Detmax1974}. However, there are two main drawbacks of DETMAX. Firstly, DETMAX does not guarantee D-optimality~\cite{Detmax1974}. Secondly, DETMAX requires numerous runs to avoid a local optimum by randomizing the initial guess. Despite these drawbacks, the DETMAX algorithm is still considered as a reference thanks to its robustness and simplicity to find reasonably good solutions. Daney et al.~\cite{Daney2005} took DETMAX algorithm further by combining it with a Tabu search algorithm which has reduced the sensitivity to local minimum as well as increased robustness to measurement noise compared to the original DETMAX. However, its efficiency was only validated with a 6 DoF Gough platform. Nevertheless, these exhaustive searching algorithms may find sub-optimal sets of postures and rely largely on randomization of the initial guess. 
    Interestingly, Kamali et al.~\cite{Kamali2019} formulated the OCP generation problem as a semi-positive convex optimization problem. They first assigned binary weights to each posture, indicating their selection, before relaxing the combinatorial problem by using continuous weights and imposing a condition that their values must be greater than one. Finally, the authors argued that a sub-optimal solution could be achieved if the pool was much larger than the number of selected postures. Still, the number of selected postures was rather arbitrarily determined even though it outperformed random selection of the same number of postures.

Previous studies commonly selected observability indices and alphabetical optimality for choosing OCP~\cite{Sun2008, JOUBAIR2013_compare_index}. These studies found that when properly scaled, all indices are equivalent in their ability to identify OCPs. However, the observability index $O_1$ (defined in Section IV.B) was highlighted as the most suitable criterion for its scaling invariance property. 
    %Criterion $O_1$ maximizes the root of the product of the singular values  of the Jacobian \cite{BormMenq1991}, with $m$ being the dimension of dataset as follows:
     %       \begin{equation}
     %       O_1 = \frac{(\sigma_1  \sigma_2  ...  \sigma_{N_B})^{1/{N_B}}}{m}
     %       \end{equation}
    
This index, akin to the D-optimality criterion, represents the geometric mean of the singular values of the full-rank Jacobian matrix of the calibration model divided by the number of samples~\cite{Detmax1974}. Given the importance of robustness to parameter scaling for humanoids, there is nowadays a consensus to retain the $O_1$ criterion \cite{bonnet2023}.

%\textcolor{blue}{
 %   \subsubsection{Observation indices for selecting OCP}
%    In geometric calibration, observation indices are used for selecting optimal exciting posture or motion design problems. These indices are calculated using the singular values $\boldsymbol{\sigma}(\mathbf{R_B}) = [\sigma_1, ... \sigma_{n_B}]$ of the full-rank Jacobian matrix as follows.
%        \begin{enumerate}
%            \item Criterion $O_1$, maximizing the root of the product of the singular values  of the Jacobian \cite{BormMenq1991}, with $m$ is the dimension of dataset :
%            \begin{equation}
%            O_1 = \frac{(\sigma_1  \sigma_2  ...  \sigma_{N_B})^{1/{N_B}}}{m}
%            \end{equation}
%            \item Criterion $O_2$, maximizing the inverse condition number \cite{Driels1990}:
%            \begin{equation}
%            O_2 = \frac{min(\sigma_1,  \sigma_2,  ...,  \sigma_{N_B})}{max(\sigma_1,  \sigma_2,  ...,  \sigma_{N_B})}
%            \end{equation}
%            \item Criterion $O_3$, maximizing the minimum singular value \cite{Nahvi1994}:
%            \begin{equation}
%            O_3 = min(\sigma_1,  \sigma_2,  ...,  \sigma_{N_B})
%            \end{equation}
%        \end{enumerate}
%}

\section{Methods}\label{sec:methodology}

\subsection{Calibration model with planar constraints}
    \subsubsection{Calibration model}
     
    \begin{figure}[ht!]\centering
      \includegraphics[width=0.95\linewidth]{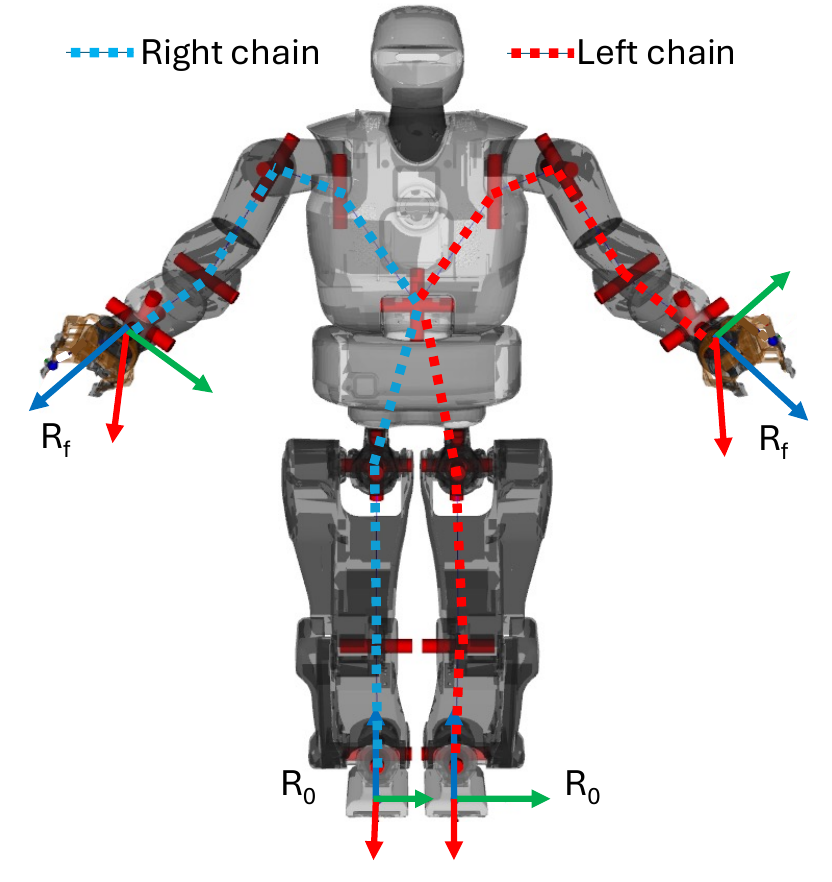}
        \caption{ Representation of 32 DoF of the TALOS humanoid robot and sub-kinematics rigth and left chains that are considered separately.}\label{fig:TALOS_structure}
    \end{figure}
    
    Fig. \ref{fig:TALOS_structure}. depicts the large size humanoid robot TALOS (175cm, 95kg). The structure of its kinematic tree consists of 32 revolute joints distributed into 5 branches. There are 2 torso joints, 6 joints for each leg, 7 joints plus a gripper joint for each arm and 2 joints that move the head of the robot. Nowadays, it is common to describe humanoid robots using URDF convention as it is used by ROS applications. The URDF convention for geometric calibration has been rarely used in the literature. This is  due to the fact that URDF convention is redundant. Indeed, it describes the transformation between two consecutive joints with a 6 dimensional vector  of the nominal geometrical parameters $\mathbf{X}_i=[{p_x}_i, {p_y}_i, {p_z}_i, {\phi_x}_i, {\phi_y}_i, {\phi_z}_i]^T$. It is well known that not all these parameters can be identified. Denavit and Hartenberg~\cite{DHparameters} showed that four or five parameters, depending on the kinematic structures, are required to fully describe the transformation between consecutive joints. 
    However, as described in Section~\ref{subsec:identifiability}, the identifiability of the geometric parameters can be dealt with using a numerical approach by setting automatically non observable parameters to zero. Then, the actual number of parameters used in URDF will be similar to the one used in Denavit and Hartenberg. Therefore, it is of a great convenience to use URDF convention for geometric calibration and kinematic calculations.

    In this paper as illustrated in Fig. \ref{fig:TALOS_structure}, two kinematic chains, left and right, of $N_J=15$ joints, going from the model base frame $R_0$, attached to the foot, and to the robot flange frame $R_{f}$, attached to the wrist of the same side, were analyzed separately. 
    The forward kinematics model allows to calculate the 3D position and orientation of the end-effector in its corresponding foot frame with respect to the joint configuration vector $\mathbf{q}\in\mathbb{R}^{N_J}$ and a vector $\mathbf{\Delta X}\in\mathbb{R}^{(N_J+2) \times 6}$ of the unknown variation of the geometric parameters, in which the sub-vector  $\mathbf{\Delta X}_i=[{\Delta p_{x}}_i, {\Delta p_{y}}_i, {\Delta p_{z}}_i, {\Delta\phi_{x}}_i, {\Delta\phi_{y}}_i, {\Delta\phi_{z}}_i]$ representing the variation of the $i-th$ joint placement, and two other sub-vectors of $6$ parameters describing the transformations from $R_0$ to the first joint frame of the kinematic chain, and from its corresponding last joint frame to $R_{f}$.
    
    %Converting the homogeneous transformation matrices $^{base}{\mathbf{T}_{EE}}$ to position and Euler angles orientation gives  the vector $\mathbf{P}_{EE}\in\mathbb{R}^{N_m}$, $N_m$ being the number of measurements, which is then a function of joint configurations and geometric parameters.
    %\begin{equation} \label{eq:FKM_function}
    %    \mathbf{P}_{EE}    \\ = \text{FKM}({\mathbf{q}}, {\mathbf{p}}+{\mathbf{\Delta p}})
    %\end{equation}
        
   % \noindent where    $\mathbf{\Delta p}=[\mathbf{\Delta p}_{1},\cdots,\mathbf{\Delta p}_{N_J+12}]$. $\mathbf{\Phi}_{g\,j}=[\delta p_{x\,j}, \delta p_{y\,j}, \delta p_{z\,j}, \delta\phi_{x\,j}, \delta\phi_{y\,j}, \delta\phi_{z\,j}]$  was the variation vector of geometric parameters that is to be identified.  
    \begin{figure}[ht]\centering
        \includegraphics[width=0.8\linewidth]{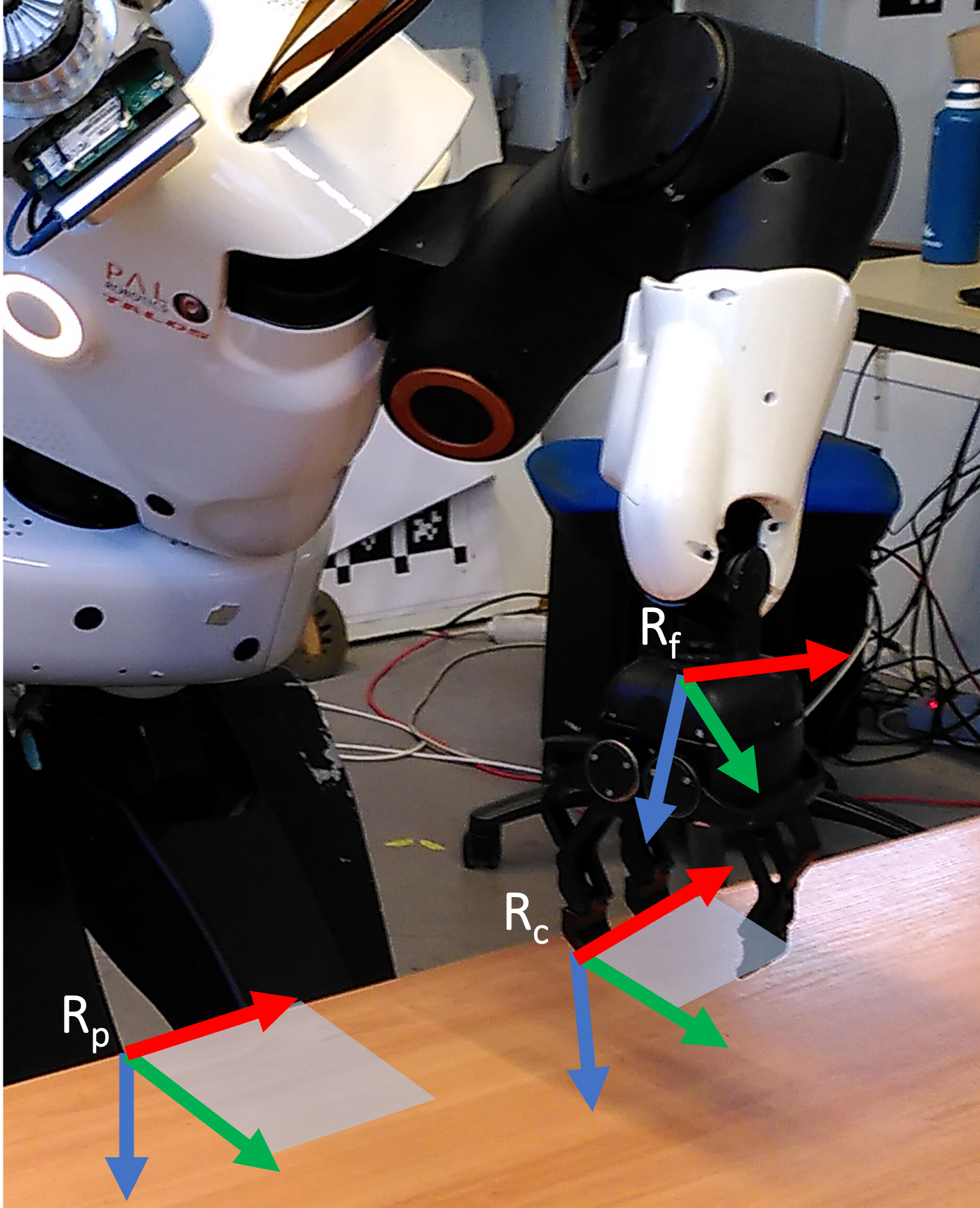}
        \caption{ Description of the flange, contact and plane frames used to calibrate TALOS robot using a plane.}\label{fig:contact_zoomin}
    \end{figure}
    
    This study introduces a novel approach for plane-constrained calibration, which differs from previous methods that generally utilize a single contact point. The key idea is to create a planar constraint by establishing simultaneous contact between three fingers of the rigid robot gripper and a flat table. The plane $R_p$ and contact $R_c$ frames are exemplified in the Fig. \ref{fig:contact_zoomin}.

Let $R_p=(O_P, x_P, y_P, z_P)$ be a frame attached to the plane to which the robot will come in contact and called the \textit{plane frame}, such that the $z$-axis is normal to the plane.
Let $R_c=(O_C, x_C, y_C, z_C)$ be a frame fixed to the wrist such that the 3 contact points belong to the $(O_C, x_C, y_C)$ plane and $O_C$ belongs to the $z$-axis of the wrist joint. The plane parameters are gathered in the vector $$\boldsymbol{\planeParameters} = [^f\mathbf{z}_{c},^f\mathbf{\boldsymbol{\phi}_x}_{c},^f\mathbf{\boldsymbol{\theta}_y}_{c},^0\mathbf{z}_{p},^0\mathbf{\boldsymbol{\phi}_x}_{p},^0\mathbf{\boldsymbol{\theta}_y}_{p}]^T$$ and are unknown, thus they should be identified. The parameters are defined as follows:
\begin{itemize}
\item $^0\mathbf{z}_{p}$: the $z$ coordinate of $O_P$ in the world frame $R_0$,
\item $^0\mathbf{\boldsymbol{\phi}_x}_{p}$: the roll angle of the contact plane with respect to the world frame,
\item $^0\mathbf{\boldsymbol{\theta}_y}_{p}$: the pitch angle of the contact plane  with respect to the world frame.
\end{itemize}
\begin{itemize}
\item $^f\mathbf{z}_{c}$: the $z$ coordinate of $O_c$ in the frame of the wrist joint,
\item $^f\mathbf{\boldsymbol{\phi}_x}_{c}$: the roll angle of the contact frame with respect to the wrist joint frame,
\item $^f\mathbf{\boldsymbol{\theta}_y}_{c}$: the pitch angle of the contact frame  with respect to the wrist joint frame.
\end{itemize}

When a contact is made at a given posture, the relative partial pose between $R_p$ and $R_c$ is constrained, providing a measurement as:
\begin{equation}
    \left[\begin{array}{l}
  ^c\mathbf{z}_{p}\\
  ^c\mathbf{\boldsymbol{\phi}_x}_{p}\\
  ^c\mathbf{\boldsymbol{\theta}_y}_{p}
\end{array}\right]= 0
\end{equation}
where $^c\mathbf{z}_{p}$ is the $z$ coordinate of $O_C$ in $R_p$, $^c\mathbf{\boldsymbol{\phi}_x}_{p}$ and $^c\mathbf{\boldsymbol{\theta}_y}_{p}$ are the roll and pitch angles of $R_c$ in $R_p$.
The estimate of that relative partial pose is a function of the joint configuration vector $\mathbf{q}_i$ and the constant and unknown vectors $\mathbf{\Delta X}$ and $\dplaneParameters=\planeParameters-\planeParameters_0$ (the variation of the plane parameters with respect to their nominal values). For clarity, we drop the dependency on nominal geometric parameters $\mathbf{X}$ and $\planeParameters$.

\begin{equation}\label{eq:estimate}
    f(\mathbf{q}_i, \mathbf{\Delta X}, \dplaneParameters) =\left[\begin{array}{l}
  ^c\mathbf{\hat{z}}_{p}\\
  ^c\mathbf{\boldsymbol{\hat{\phi}}_x}_{p}\\
  ^c\mathbf{\boldsymbol{\hat{\theta}}_y}_{p}
\end{array}\right]
\end{equation}
% where $^f\mathbf{z}_{c}$, $^f\mathbf{\boldsymbol{\phi}_x}_{c}$, $^f\mathbf{\boldsymbol{\theta}_y}_{c}$ were the  plane parameters expressed in $R_f$  thanks to the forward kinematics model. 

Thus, for $N$ postures, the  measurement functions are stacked into the following measurement vector:
\begin{equation}\label{eq:measurements}
\mathbf{\hat{Y}}(\mathbf{\Delta X}, \dplaneParameters)=\left[\begin{array}{c}
f(\mathbf{q}_1, \mathbf{\Delta X}, \dplaneParameters)\\
\vdots\\
f(\mathbf{q}_N, \mathbf{\Delta X}, \dplaneParameters)
\end{array}\right]
\end{equation}

Since the measurements are zero, the calibration problem boils down to minimizing the sum of squares of estimates of the relative partial pose between $R_p$ and $R_c$: 

\begin{equation}\label{eq:calibration_model}
\begin{aligned}
\text{Find} \ \ [\mathbf{\Delta X}^* \dplaneParameters^*]\ \text{solution of} \ \ \underset{[\mathbf{\Delta X}  \dplaneParameters ]}{\text{min}} \
\lVert\mathbf{\hat{Y}}(\mathbf{\Delta X}, \dplaneParameters)\rVert^2
\end{aligned}
\end{equation}

    \subsubsection{Identifiability of geometric parameters} \label{subsec:identifiability}
    
    % All six geometric parameters for each joint of a kinematic chain cannot be identified ~\cite{khalil_gautier_enguehard_1991}. The maximum number of identifiable parameters for given kinematic structure is at most only four out of six  geometric parameters for a revolute joint, and 2 out 6 geometric parameters for prismatic joint. 
    % Furthermore, Everett et al.~\cite{Everett1988} also showed that the identifiability of these parameters is also influenced by the placement of joints in the kinematic tree. For example, if two consecutive joints are co-linear it is impossible for both orientationnal parameters, i.e. along joint axes, to be identified separately as they are dependent.
    % Parameters dependency is observable in the Jacobian matrix, relating the geometric parameters to be identified with the measurements, by analysing its rank. The set of identifiable parameters can be determined \textit{a priori} using random joint configurations and by calculating the QR decomposition of the resultant Jacobian matrix~\cite{khalil_gautier_enguehard_1991}. The non influencial parameters can then be removed from the Jacobian matrix and the linearly dependant ones can be grouped together~\cite{Khalil_chapter11}.
    % In a physical robotic system characterized by the interconnection of rigid bodies through mechanical joints featuring a constrained number of Degrees of Freedom (DoF), the identification of all six geometric parameters for each joint within a kinematic chain is unattainable~\cite{khalil_gautier_enguehard_1991}. 
    The maximum number of identifiable parameters varies depending on the kinematic structure. In the case of a revolute joint, only a maximum of four out of the six geometric parameters can be identified, while for a prismatic joint, only two out of the six parameters can be identified~\cite{Everett1988}. Additionally, the identifiability of these parameters is influenced by the arrangement of joints in the kinematic tree, as highlighted by Everett et al.~\cite{Everett1988} and Meggiolaro et al.~\cite{Meggiolaro2000}. If two consecutive joints are co-linear or parallel, it becomes infeasible to identify both orientation parameters separately since they are interdependent. The interdependence of parameters can be observed by numerically analyzing the rank of the Jacobian matrix, which relates the geometric parameters to be identifiable with the measurements. In the case of the TALOS robot, each considered kinematic chain consists of 15 revolute joints, in which the first torso joint and the first leg joint of each leg are co-linear. %Lastly, the type of measurements and the quality of data set obtained from the 3-point contact plane-constrained calibration prevents the identifiability from being diminished due to the technical limitation.

    We denote by $\mathbf{R}$ the Jacobian matrix of $\mathbf{\hat{Y}}$.
    To determine the set of identifiable parameters, the QR decomposition of $\mathbf{R}$ can be calculated based on random joint configurations~\cite{khalil_gautier_enguehard_1991}. Non-influential parameters can then be eliminated from the Jacobian matrix, and linearly dependent parameters can be grouped together in linear combinations with independent parameters. Note that the physical interpretation of the linear combinations has been discussed in~\cite{Meggiolaro2000}.
    By performing QR decomposition on the matrix $\mathbf{R}$,
    a full-rank base regressor matrix $\mathbf{R}_{b}\in\mathbb{R}^{3N \times N_b}$ and \textit{so-called} base geometric parameters $\mathbf{\Delta X}_{b}\in\mathbb{R}^{N_b}$ can be obtained to present the final expression of plane-constraint calibration model as follow:
        \begin{equation} \label{eq:base_regression}
        \mathbf{\Delta\hat{Y}} = \mathbf{R}_{b} \mathbf{\Delta X}_{b} 
        \end{equation}
        Note that base geometric parameters in $\mathbf{\Delta X}_{b}$ are linear combinations of geometric parameters in $\mathbf{\Delta X}$.
        $$
        \mathbf{\Delta X}_{b} = \mathbf{A}\ [\mathbf{\Delta X}, \dplaneParameters]
        $$
        where $\mathbf{A}$ is a sparse $N_b\times N_p$ constant matrix.
        We compute vector $\mathbf{\Delta X}_{b}$ by minimizing the squared norm of $\Delta \mathbf{Y}$ using Levenberg-Marquard algorithm. 
        % \textcolor{red}{EXPLAIN HOW WE FIND STANDARD PARAM. we need to add how we determine delta X from delta Xb}
        %We then get the parameters to identify by minimizing the deviation to the nominal model using the pseudo-inverse:
        %$$
        %(\mathbf{\Phi}^T (\mathbf{\Phi}_1 - \mathbf{\Phi}_1^0)^T)^T = A^{+}\ \mathbf{\Phi}_{gb}
        %$$
    % \textcolor{red}{The number of parameters need to be better explained please that add some more info. what is the value of $N_b$ at the end ?}    
    In summary, on each kinematic chain, there are 57 identifiable parameters in which 4 parameters located at torso joints are mutual, resulting in $N_b = 110$ parameters in total to be identified for the robot. %\textcolor{blue}{In fact, the numbers of identifiable parameters are aligned with the analytical method of eliminating redundant parameters in robot calibration~\cite{Meggiolaro2000}. }
    
    There are a few techniques to re-map base parameters $\mathbf{\Delta X}_{b}$ back to $\mathbf{\Delta X}$ in order for matching the URDF format. However, all of these techniques would never give a unique result for $\mathbf{\Delta X}$. Therefore, for the sake of simplicity without losing any accuracy on the calibration process, the re-mapping is done by setting linearly dependent parameters to zero while setting the independent parameters equal to the corresponding base parameter value.

\subsection{Information ranking algorithm for selecting optimal calibration postures}\label{section:iroc}

    A key challenge in the calibration process is selecting an optimal set of calibration postures that provide maximum information about the robot's geometric parameters while minimizing the number of measurements required.
    As discussed in Section \ref{subsection:select_ocp}, traditional methods for selecting calibration postures, such as DETMAX \cite{Detmax1974}, often struggle with the high dimensionality of humanoid robots and can be computationally expensive. Furthermore, they may not guarantee global optimality or automatically determine the optimal number of postures. To address these limitations, we propose an Information Ranking algorithm for selecting Optimal Calibration postures (IROC).
    This novel method not only offers a practical way to select optimal postures based on ranking, but also provide a mean to determine the necessary number of optimal postures.

    % \subsection{Problem formulation and observability criteria}
            
    % To evaluate the information content of different postures and select optimal calibration configurations, we need to define appropriate criteria and formulate the problem mathematically. This formulation serves as the foundation for our novel IROC algorithm.
    
    \subsubsection{Observability criterion selection}
    
    The information matrix $\mathbf{I}_B = \mathbf{R}_b(\mathbf{q})^\textbf{T} \mathbf{R}_b(\mathbf{q})$ can be computed over a set of configurations $\mathbf{q}$. It represents the amount of information the regression model possesses about the parameters to be identified. As such, it serves as a tool to infer the optimality of a dataset. Other studies \cite{Hollerbach2008} have interpreted observability indices as aspects of a hyper-ellipsoid in $\mathbf{\Delta\hat{Y}}$ space (Eq. \eqref{eq:base_regression}), with the parameters vector $\mathbf{\Delta X}_{b}$ as a hyper-sphere. As previously discussed, the most popular index, $O_1$, maximizes the volume of this hyper-ellipsoid and thus is less sensitive to parameters scaling problems. $O_1$ was then retained in this study as was defined as follows:

    \begin{equation}
    O_1(\mathbf{q}) = \frac{(\det{\mathbf{I}_B(\mathbf{q}))}^{\frac{1}{2N_b}}}{\sqrt{k}}
    \label{eq:O_1 criterion}
    \end{equation}
    
    where $N_b$ is the number of base parameters and $k$ is the number of considered postures. Each posture independently contributes to the information matrix, allowing $O_1$ to be computed as the element-wise sum of individual information matrices.

    %while $O_2$ minimizes eccentricity, making the hyper-ellipsoid closer to a hyper-sphere. $O_3$ maximizes the length of the shortest axis, improving the worst-case scenario.
    
   % Among these, the observability index $O_1$ is determined to be the most suitable criterion for scaling invariance. Given the importance of robustness to parameter scaling in high-DOF humanoid robot calibration, we choose the $O_1$ criterion for selecting optimal calibration postures in this study.
    
    \subsubsection{General problem formulation for selecting optimal postures}

    The problem of finding an optimal set of postures, $\mathbf{q}$, that maximizes $O_1$ can then be formulated as the following combinatorial optimization problem:
    
    \begin{equation}
    \label{optim_zmp}
    \max_{\mathbf{q} \subset \mathbf{Q}} \quad O_1(\mathbf{q})
    \end{equation}

    where $\mathbf{Q}$ denote a pool of $N$ feasible candidate configurations sampled in a simulated environment. 
    
    \subsubsection{IROC: A novel approach to optimal posture selection}
    
    The problem of selecting optimal calibration postures presents significant computational challenges. A straightforward approach would be to phrase it as a mixed-integer optimization problem, where binary weights (1 for inclusion, 0 for exclusion) are assigned to each candidate posture, subject to a constraint on the total number of selected postures. The objective would be to maximize the information content of the selected postures. However, solving such a mixed-integer optimization problem at a large scale is computationally prohibitive, especially for high-DOF systems like humanoid robots.
    
    The proposed IROC algorithm offers several advantages over traditional methods such as: reducing computational burden, faster convergence while maintaining optimality, and automatic selection of the minimum number of optimal calibration postures.
    
     IROC operates as follows:
    
    \begin{itemize}
        \item Continuous weight assignment: Instead of binary weights, we assign a random positive continuous weight to each candidate posture's information matrix, as suggested by Kamali et al.~\cite{Kamali2019}. These weights are normalized to sum to 1, forming a constraint for the subsequent optimization problem.
        \item Information-based optimization: We formulate an optimization problem to determine the weight values that maximize the determinant of the weighted sum of information matrices. This approach intuitively assigns higher weights to postures that contain more information.
        \item Ranking and selection: By sorting the optimized weights in a descending order, a ranking of candidate postures based on their information content can be obtained. Then, the highest-ranked postures, ie the ones that improve the $O_1$ criterion, can be selected efficiently.
    \end{itemize}
    
    The detailed steps of the IROC algorithm are presented in Algorithm~\ref{algo:proposed_algo}.
    
    \begin{algorithm}[!ht]
    \caption{IROC algorithm }
    \begin{algorithmic}
    
        \Require $\mathbf{Q} \gets$ \text{ a set of $N$ candidate postures}
        
        \State $\mathbf{\Omega} = (\omega_1, ..., \omega_{N}) \gets$ \text{a vector of $N$ weighting variables}
        
        \State $\sum_{i=1}^{N} \omega_i = 1$
        
        \State $\mathbf{q}_i \in \mathbf{Q}: {\mathbf{I}_B}_i(\mathbf{q}_i) \gets$ \text{information matrix by the posture $\mathbf{q}_i$}
        
        \Ensure $\mathbf{q^{*}}$
        \State $\mathbf{\Omega^{*}} = \text{argmax}
        \frac{(\det{(\sum_{i=1}^{N} \omega_i \mathbf{I}_B}_i(\mathbf{q}_i)))^{\frac{1}{2N_b}}}{\sqrt{N}}$
        \\
        \State $(\mathbf{\Omega^{*}}_{des}, \mathbf{Q}_{des}) \gets $Descending sort $(\mathbf{\Omega^{*}}, \mathbf{Q})$
        
        \State $k = k_0$ \text{, $k_0 = 21$: required minimum number of meas.}
        \While {$\mathbf{q^{*}} = [ \ ] \ \mathbf{and} \ k \leq N$}
            \State $\mathbf{q_k} = \gets {\mathbf{Q}}_{des}\left[1:k\right] $
            \State $ {O_1}^k = \frac{(\det{\mathbf{I}_B(\mathbf{q_k}))}^{\frac{1}{2N_b}}}{\sqrt{k}} $
            \If {${O_1}^{k} - {O_1}^{k-1} \leq 0$}
                \State $\mathbf{q^{*}} \gets \mathbf{q_{k-1}}$
            \EndIf
            \State $ k = k + 1$
        \EndWhile
        \end{algorithmic}
        \label{algo:proposed_algo}
    \end{algorithm}
    
    In the following sections, we will discuss the generation of candidate postures, compare IROC with traditional methods, and analyze its performance in detail.

    \subsubsection{Generation of candidate postures}
     %   \begin{figure}[ht!]\centering
     %       \includegraphics[width=0.95\columnwidth]{figures/talos_contact_model.png}
     %       \caption{Location of the 18 contact points, used to generate random optimal postures, located on a planar platform.} \label{fig:talos_contact_setup}
     %   \end{figure}

        Using HPP library \cite{Mirabel2016hpp}, a pool of 1500 random and feasible postures were the robot is in contact with a table at 18 different locations was generated. To do so, as the task is redundant, first, a  random joint configuration following a uniform distribution in the joint limits was generated.
        Second, the quasi-static whole-body equilibrium and gripper-plane contact constraints were ensured using a Gauss-Newton algorithm initialized with the random configuration. Finally, collisions and joint limits were checked.

%\endgroup
%%%%%%%%%%%%%%%%%%%%%%%%%%%%%%%%%%%%%%%%%%%%%%%%%%%%%%%%%%%%%

\section{Results}\label{sec:results}
%\begingroup
%\color{red}
%\subsection{Optimal calibration postures selection}
%    \begin{figure}[ht!]\centering
%        \includegraphics[width=0.9\linewidth]{figures/TALOS_1500samples_fixed.png}
%        \caption{ (a) $O_1$ criterion evolution calculated by iteratively stacking postures, (b) Sorted weight values of each information matrix using IROC.} \label{fig:SOCP_simul}
%    \end{figure}
%    Fig.~\ref{fig:SOCP_simul} shows the resulting performance of the IROC algorithm, presented in Algorithm~\ref{algo:proposed_algo}, in selecting the optimal postures. When the entire pool of 1500 random postures was used, the maximal value of the $O_1$ criterion was $1.6$. As depicted in Fig.~\ref{fig:SOCP_simul}.a, this value can be reached with as few as $41$ postures when they are ranked optimally using IROC. Fig.~\ref{fig:SOCP_simul}.b. displays the weight attribution of each posture, where each weight reflects the level and quality of information contained in each posture. One can observe, by looking at the abrupt drop in weight values, that after $212$ postures, additional postures bring negligible information to the calibration process.

%\endgroup
%%%%%%%%%%%%%%%%%%%%%%%%%%%%%%%%%%%%%%%%%%%%%%%%%%%%%%%%%%%%%

      \subsection{Performance analysis of IROC}

  \begin{figure}[ht!]\centering
            \includegraphics[width=1\columnwidth]{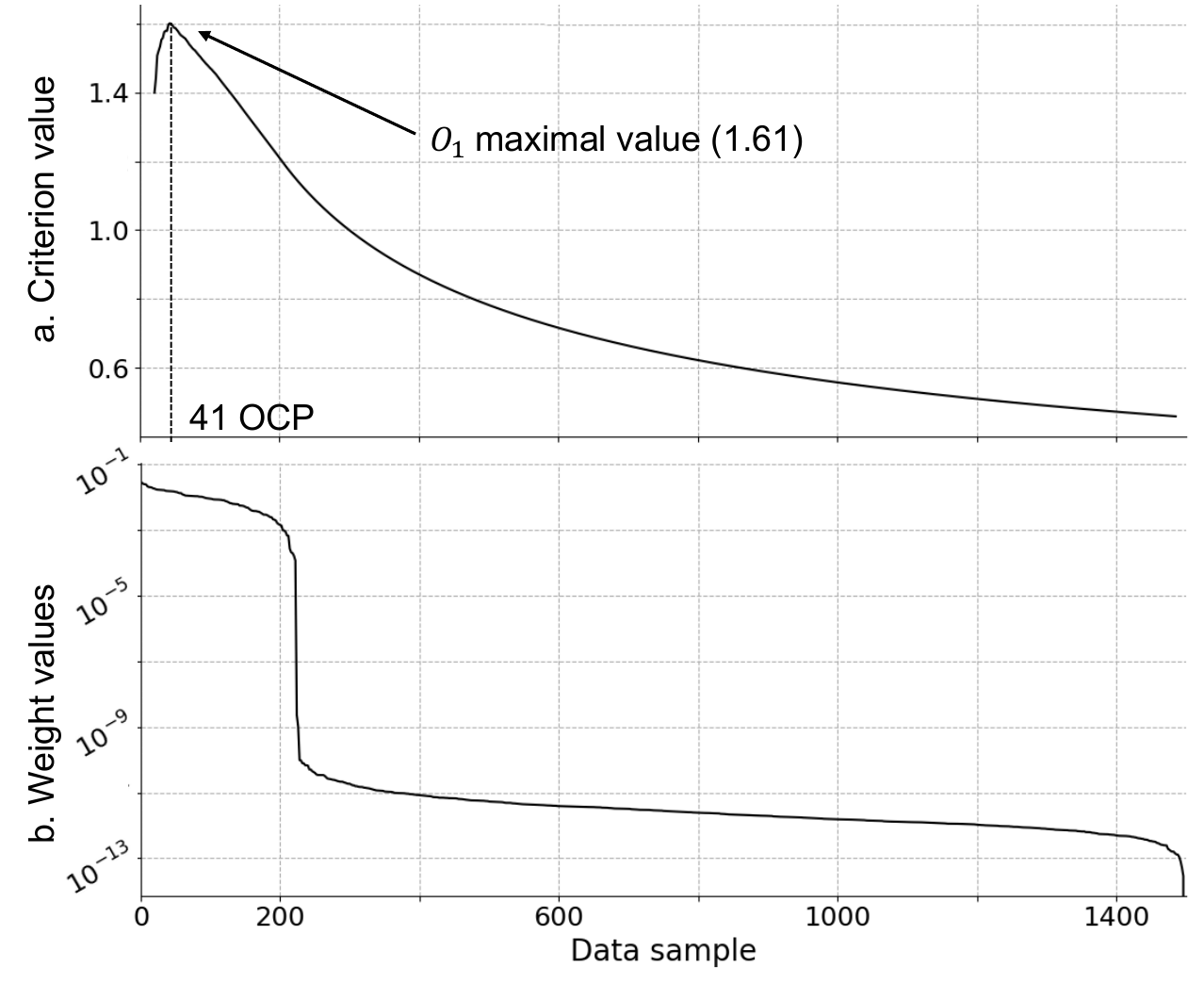}
            \caption{  $O_1$ criterion evolution calculated by iteratively stacking postures (a). Sorted weight values of each information matrix using IROC (b) .} \label{fig:SOCP_simul}
        \end{figure}
        Figure~\ref{fig:SOCP_simul} shows the resulting performance of the IROC algorithm in selecting the optimal postures. When the entire pool of 1500 random postures was used, the maximal value of the $O_1$ criterion was $1.61$. As depicted in Fig.~\ref{fig:SOCP_simul}.a, this value can be reached with as few as $41$ postures when they are ranked optimally using IROC. This demonstrates the algorithm's ability to efficiently identify a minimal set of highly informative postures.
        
        Fig.~\ref{fig:SOCP_simul}.b displays the weight attribution of each posture, where each weight reflects the level and quality of information contained in that posture. One can observe, by looking at the abrupt drop in weight values, that after $212$ postures, additional postures bring negligible information to the calibration process. This feature of IROC allows for automatic determination of the optimal number of calibration postures, a capability not present in methods like DETMAX.

      \begin{figure}[ht!]\centering
        \includegraphics[width=\columnwidth]{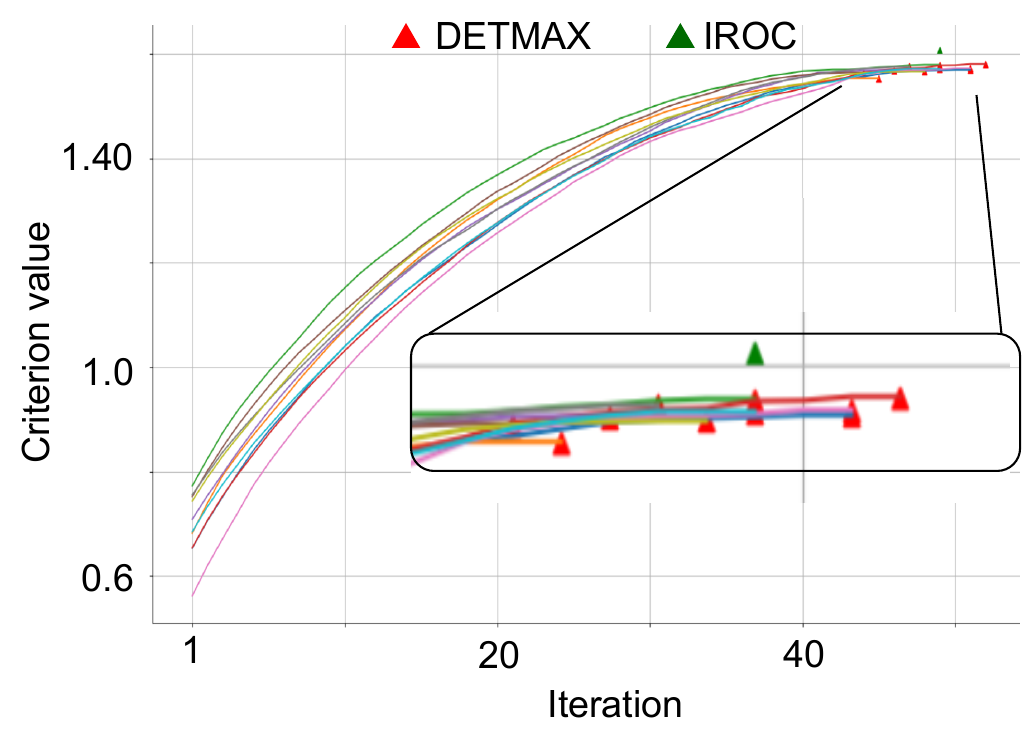}
        \caption{ Comparison of IROC (green triangle) and of the DETMAX algorithms (red triangles) for $O_1$ maximization. DETMAX was run 10 times as indicated by color lines. } \label{fig:detmax_iroc_compare}
    \end{figure}
    
    To highlight the advantages of IROC, it was compared with the widely-used DETMAX method. Implementation details of DETMAX can be found in \cite{Detmax1974}. DETMAX main limitations compared to IROC are that:
    \begin{itemize}
      
        \item it requires numerous runs to avoid local optima by randomizing the initial guess,
        \item the computational complexity increases rapidly with the number of candidate postures, 
        \item it does not automatically determine the optimal number of postures to select.
    \end{itemize}
    %, whose algorithm is detailed in Algorithm \ref{exchange_algo}.

    %In a more relaxed approach from the above algorithm, DETMAX allows the number of adding/subtracting design points to be more than 1. Additionally, a storage of failure set prevents the algorithm from going further in the direction of non-optimality. 

    As highlighted in Fig. \ref{fig:detmax_iroc_compare}, IROC addresses these limitations by providing a more efficient and robust method for selecting optimal calibration postures. Basically, IROC achieves a better criterion value (1.61) than the best of 10 runs of DETMAX (1.58). DETMAX required in average 50 iterations to determine the optimal postures while a single iteration was performed by IROC. In term of computational time, IROC reduced it by $80\%$ when compared to DETMAX leading to 12 minutes on an Intel® Xeon® Processor E5-1660 v3 workstation machine. This makes IROC particularly suitable for high-DOF systems like humanoid robots, where the number of potential postures is large and the computational cost of repeated optimizations would be prohibitive.

\subsection{Experimental setup}

    \begin{figure}[ht!]\centering
        \includegraphics[width=1\linewidth]{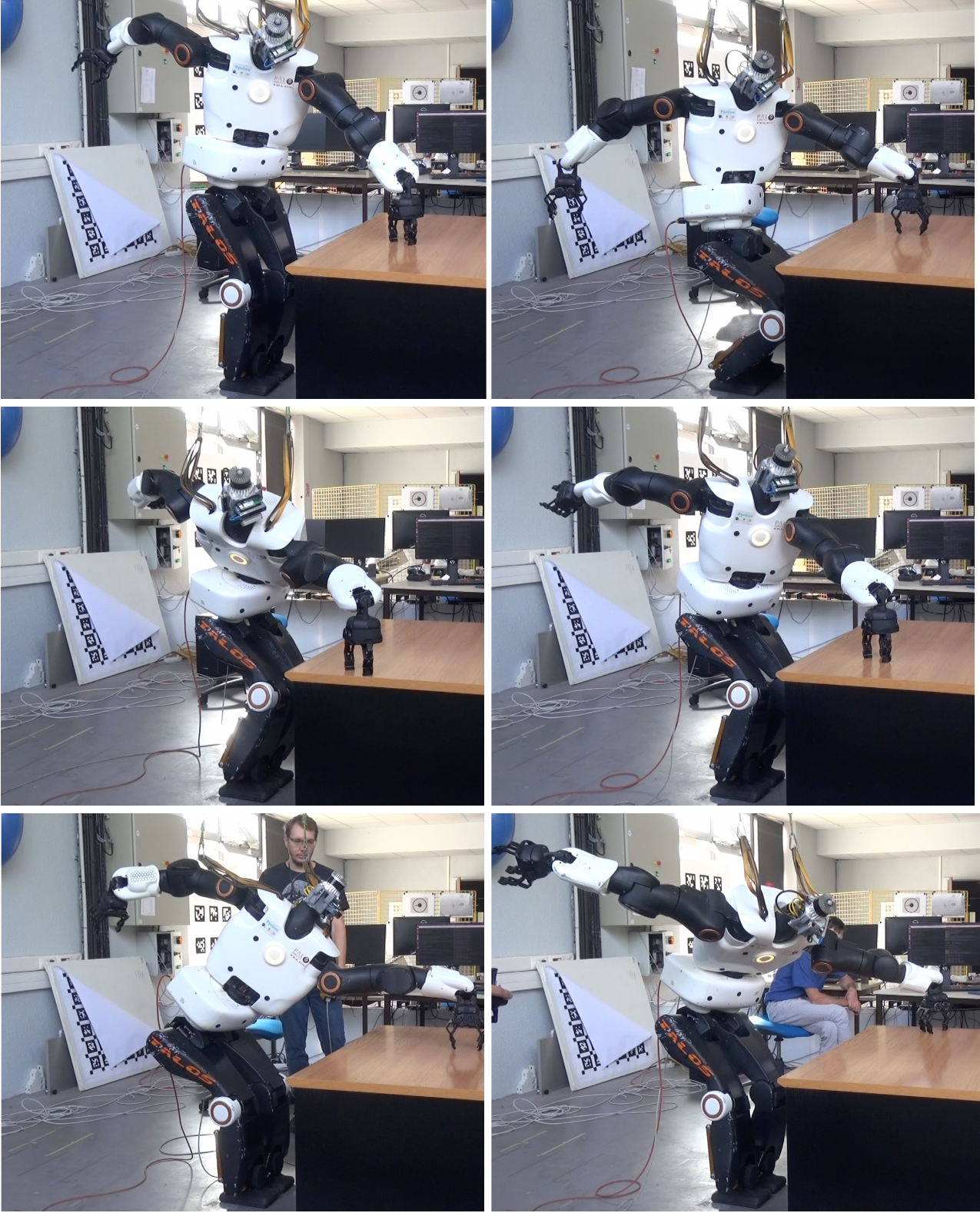}
        \caption{Experimental setup showing TALOS making 3-points contacts between its gripper and a flat table while maintaining balance and fixed feet on the ground.} 
        \label{fig:contact_1}
    \end{figure}  
    
\begin{figure}[ht!]\centering
        \includegraphics[width=1\linewidth]{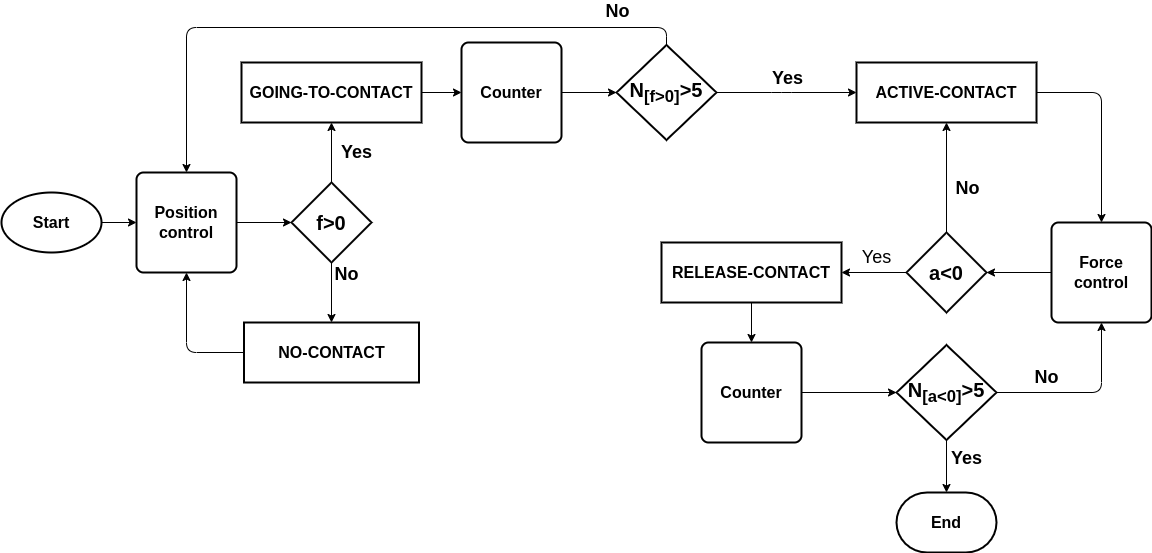}
        \caption{Control flowchart of the TALOS making contact on a table with its gripper.}\label{fig:control_flowchart}
    \end{figure}

    Fig. \ref{fig:contact_1} illustrates the experimental setup. As previously mentioned (Section~\ref{sec:related-work}) only a single plane is required to perform the calibration procedure. Thus, a flat table was positioned in front of the robot, which maintained quasi-static balance with both feet flat on the ground. The robot then made contacts on planned targets. Traveling paths connecting the postures were planned using the motion planning solvers HPP and Agimus~\cite{Mirabel2016hpp,Nicolin2020agimus}. The traveling route was optimized using a classical traveling-salesman algorithm to minimize the traveling distance. Each motion, where the robot moved from one posture to another, consisted of two sub-motions: a free motion in which the robot moved to an approaching pose just above the table, and a go-to-contact motion in which the robot transitioned from the approaching pose to make contact with the table. The first motion was executed with a position controller, and the latter was controlled using a custom admittance controller. Fig.~\ref{fig:control_flowchart} depicts the state machine diagram of the command flow for the controller. Data from a force-torque sensor located at the wrist were used to detect, maintain, and release contact. An initial contact, not necessarily distributed on all the three fingers, initiated a slow motion to go to 3-points contact.
    %$GOING\_TO\_CONTACT$ phase. 
    During this phase, if the norm $f$ of the resultant forces measured by the force-torque sensor at the wrist held above a threshold for $N_f=5$ consecutive samples, a Boolean variable would change to indicate an active contact.
    %state would change from $NO\_CONTACT$ to $ACTIVE\_CONTACT$, thus the switch between two control modes which would allow to transition from a position controller to a force controller. 
    Finally, to release the gripper from the contact mode, a custom criterion based on the relative motion between the gripper and the contact points has to be satisfied for $N_a=5$ consecutive control periods. %samples, the gripper would be considered out of the $ACTIVE\_CONTACT$ state. 
        
\begin{figure}[ht!]\centering
        \includegraphics[width=1\linewidth]{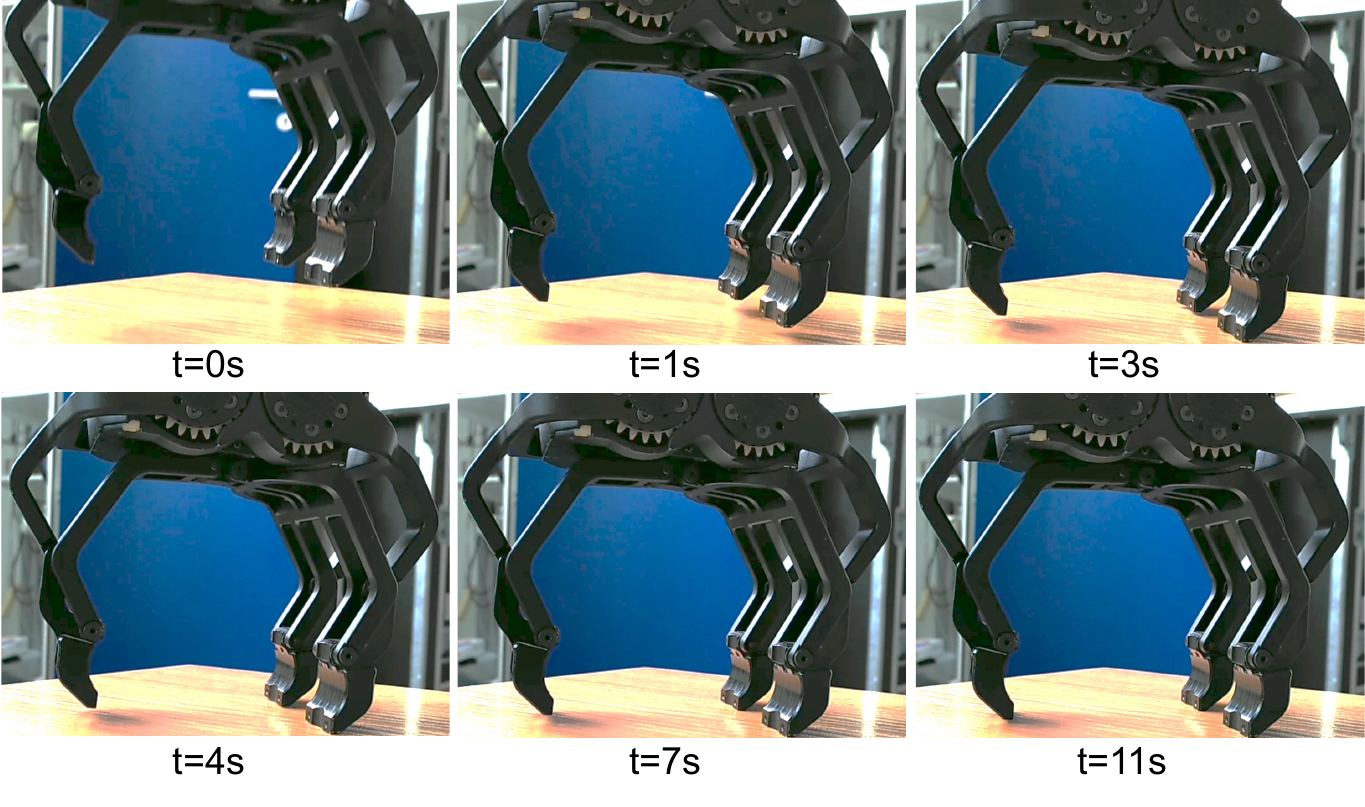}
        \caption{Example of the admittance controller action creating a 3-points contact.  } \label{fig:admittance_details}
    \end{figure}

    Planar contact was achieved and maintained by ensuring that all three fingers were in contact with the table.  To achieve this, an admittance controller was used to enforce a force normal to the plane and zero moments at the wrist. For example, it is easy to understand that contact with only one or two fingers would generate nonzero measured moments at the wrist center. In such cases, the admittance controller action would result in a rotating motion of the gripper to eliminate the undesired moments.  Fig.~\ref{fig:admittance_details} shows a motion of 11s where a three-finger plane contact was created. The cancellation of the moments was always achieved in less than 30s. Note that the fingers are made of rigid steel and thus would not be deformed.

  Once a static 3-points contact was successfully achieved, a ROS node was recording  joint encoders and force sensor data. %The hierarchy of executions was handled in real-time by the so-called software Stack of Tasks~\cite{mansard:icar:09}. 
  Overall, the experimental procedure to achieve a single planar contact was $2$ minutes on average. This is due to the fact that the transition between each postures were planned on-line. Whole-body balance was also ensured as shown in Fig. 5 where the  arm not being calibrated is used to improve balance.
  Unfortunately, our TALOS robot was not able to run continuously for more than one hour due to software issues.
  Consequently, it was not possible to perform all the $41$ best postures proposed by IROC algorithm. As it was not possible to restart the robot without modifying its feet pose, only the 31 best calibration postures were recorded and used in the calibration process. As shown in Fig. \ref{fig:SOCP_simul}.a., this will have little influence on the excitation criterion $O_1$.

\subsection{Calibration results}

    % \begin{figure*}[t!]
    %     \centering
    %         \subfigure[] 
    %         {
                
    %             \includegraphics[width=.3\textwidth]{pz_error.png}
                
    %             % .png .jpg ... according to supported graphics files
    %         }\hfill
    %         %
    %         \subfigure[] 
    %         {
                
    %             \includegraphics[width=.3\textwidth]{phix_error.png}
    %             % .png .jpg ... according to supported graphics files
    %         }\hfill
    %         %
    %         \subfigure[] 
    %         {
                
    %             \includegraphics[width=.3\textwidth]{phiy_error.png} 
                
    %         }

    %     \caption{Comparison of residual values by robot model with calibration data set before calibration (dotted-red) and after calibration (hatched-blue) in term of: (a) the absolute position along z axis; (b) the absolute orientation around x axis; (c) the absolute orientation around y axis of the robot end-effector.  }
    %     \label{fig:residual}
    % \end{figure*}
  \begin{figure}[ht!]\centering
        \includegraphics[width=1\linewidth]{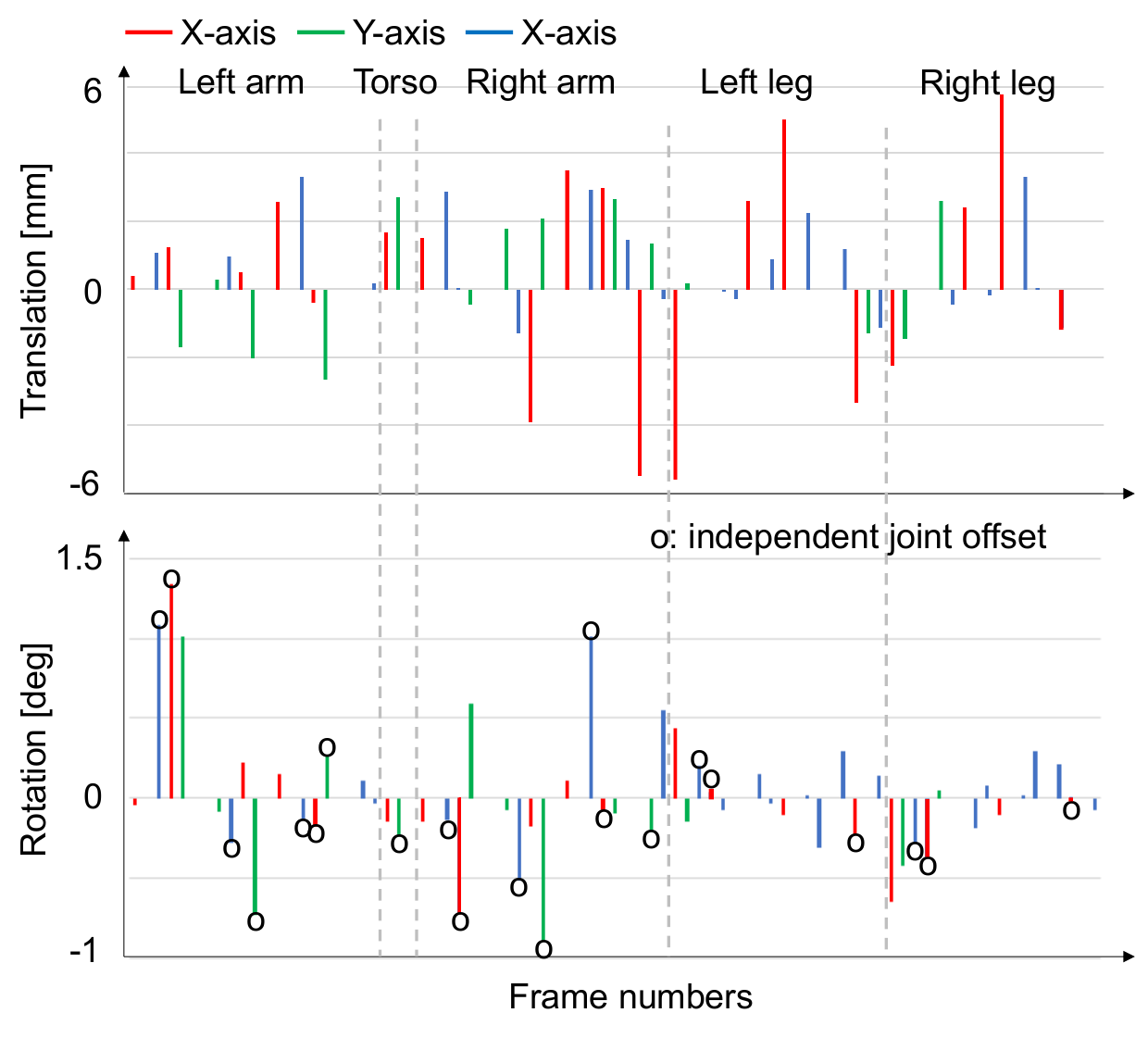}
        \caption{Values of the translational and rotational identified parameters. } \label{fig:parameters}
    \end{figure}
    
    % The RMS error values of absolute position of end-effector along z axis reduced from 37.2mm to 5.36mm. 
    % \begin{figure}[ht!]\centering
    %     \includegraphics[width=0.85\linewidth]{figures/rms_evole.png}
    %     \caption{Convergence of RMS error of the end-effector position over least-square optimization} \label{fig:rms}
    % \end{figure}
    % Fig. \ref{fig:residual} depicts the distribution of the residual errors of the calibration process, as defined in equation~(\ref{eq:measurements}) for the $31$ considered OCP. 
    After calibration, the residual error was much smaller as it was decreased in average by a factor 3. In details, the absolute errors were reduced from $9.19\pm11.64$mm to $2.88\pm3.89$mm, from $1.12\pm1.34$deg to $0.31\pm0.41$deg, and from $1.02\pm1.20$deg to $0.33\pm0.42$deg for the position error along z and for the roll and pitch angles, respectively.   
    
   Fig.~\ref{fig:parameters} shows all the identified geometric parameters. The translational parameters varied from [-6,+6]mm, and rotational ones from [-1.5,+1.5]deg. Note that 50 individual geometric parameters were set to 0 as they were not identifiable due to the kinematic structure of the robot. For example, a single joint offset could be identified for two co-linear joints. As one can see, the identified parameters are not negligible. As expected, the largest rotational parameters were the ones corresponding to the joint offsets. This is especially the case for the arms as the joint offsets of the legs were relatively small. On the contrary, some translational parameters of the leg were relatively large at the hip and knee levels. This might be explained by the fact that TALOS legs are subject to some structural flexibility that was not included in the model used in this study~\cite{bonnet2023}.
    
    An accompanying video media shows the executed motions,  partially depicted in Fig. \ref{fig:contact_1},  and the controlled go-to-contact procedure and experimental setup. 
    \begin{figure*}
     \begin{subfigure}{0.28\textwidth}
         \includegraphics[width=\linewidth]{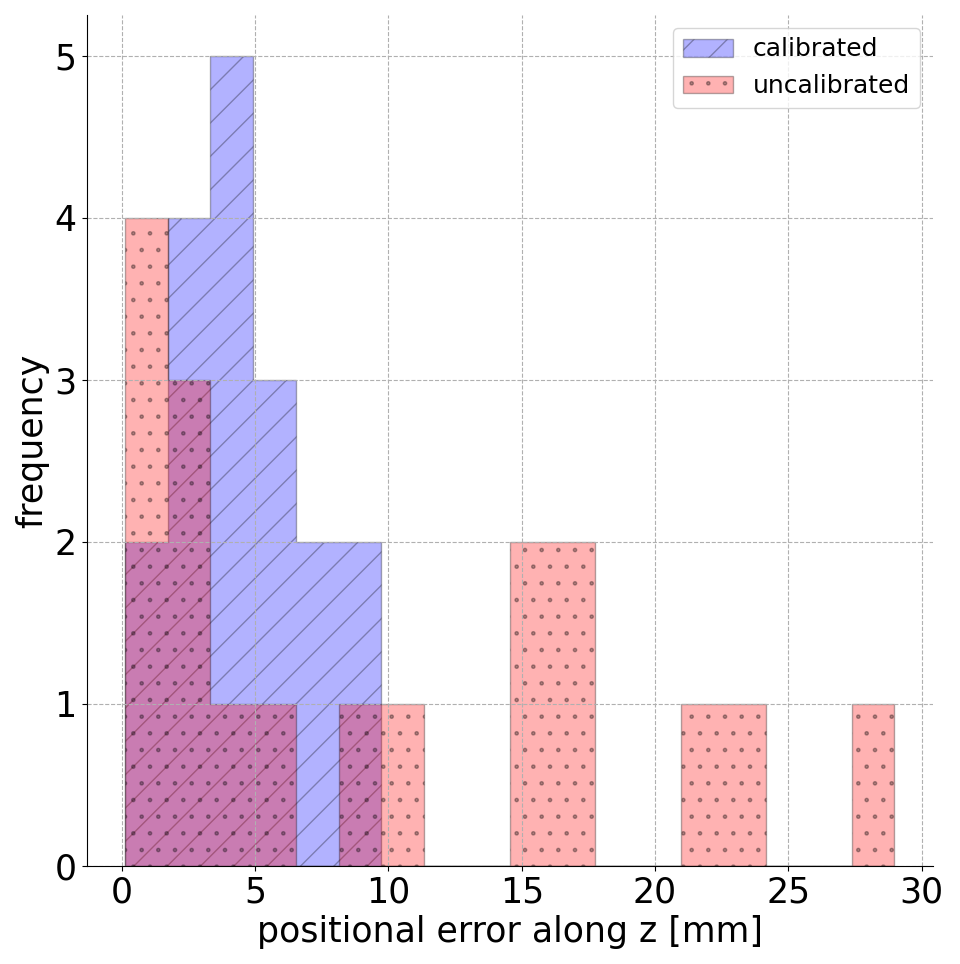}
         \caption{}
     \end{subfigure}
     \hfill
     \begin{subfigure}{0.28\textwidth}
         \includegraphics[width=\linewidth]{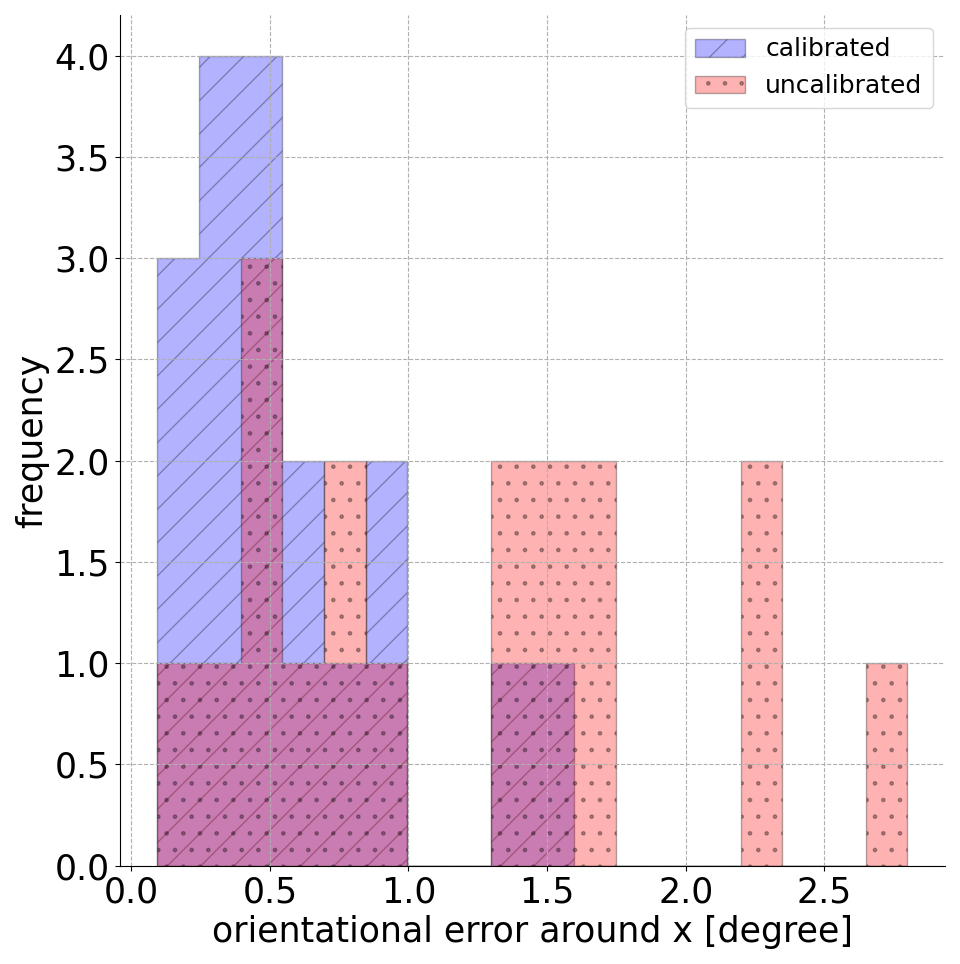}
         \caption{}
     \end{subfigure}
     \hfill
     \begin{subfigure}{0.28\textwidth}
         \includegraphics[width=\linewidth]{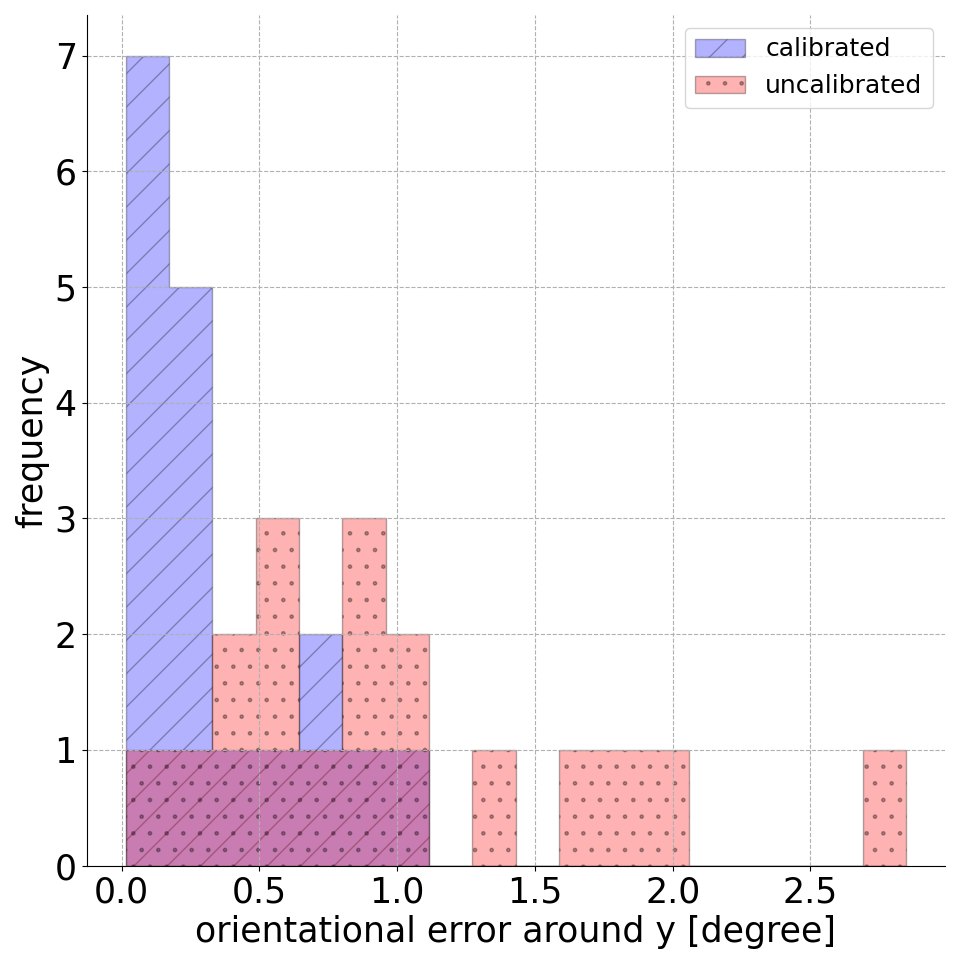}
         \caption{}
     \end{subfigure}
     
    \caption{Comparison of residual values  obtained during the experiment performed with the robot nominal model (dotted-red) and with the robot calibrated model (hatched-purple) on validation data: (a) is the absolute position along z-axis; (b) is the roll angle; (c) is pitch angle of the robot end-effector as defined in equation~(\ref{eq:estimate}).}
    \label{fig:validation}
    \end{figure*}

\subsection{Cross validation results}
    Nine random postures, which were not used in the calibration process, for each kinematic chain, were used to cross-validate the calibration of the TALOS robot. These postures were played twice on the robot: first with the nominal TALOS model and then its model was updated with the identified parameters. Doing so ensured that motion planning and control took into account the calibrated model. 
    Fig.~\ref{fig:validation} shows that the positional and orientational errors were largely reduced when using the calibrated model compared to when using the nominal model. The mean absolute error decreased by an average factor of 2.3 for positional and orientational errors. In particular, it decreased from 9.89mm to 4.54mm, from 1.18deg to 0.56deg, and from 0.98deg to 0.34deg for position error along the z-axis and orientation around roll and pitch, respectively.

\section{Conclusion}

    In this paper, we presented a new methodology for automatically estimating whole-body geometrical parameters of a large humanoid robot, specifically the TALOS, without reliance on external sensors. The proposed approach uses a simple plane and an impedance controller to establish a 3-point contact  with a flat surface. It allows to gather data for the calibration process.
    Making contacts with flat surfaces is common in manipulation and locomotion for humanoids. It is thus reasonable to develop a plane-constrained calibration method for humanoids that could be routinely used to calibrate humanoid robots.  To the best of our knowledge, it is the first time the whole-body calibration of a large-scale humanoid robot was achieved without external sensor.  

    Due to the numerous DoF of a humanoid robot and its frequent need to be re-calibrated after modification or maintenance, it is necessary to design optimal calibration postures to reduce the time of the calibration procedure. OCP, contrary to random ones, will excite the parameters to be identified best. However, generating optimal postures that guarantee mechanical limitations for such a complex system is a difficult task~\cite{bonnet2023}. In this paper, we proposed the IROC algorithm that not only can select OCP but also determine its minimal number. As shown in Fig. \ref{fig:SOCP_simul}, IROC allows analyzing the minimal number of OCP required to perform good calibration. It was convenient in our case, for example, as we could not use the 41 OCP maximizing $O_1$ criterion due to technical issues, to ensure that the 31 postures we actually used were sufficient to identify the TALOS robot correctly. The present study is limited to a pool of feasible postures of 1500, but it could be easily extended to obtain even more optimal calibration postures.
    The accuracy improvements brought by calibrated model were compared and validated using cross validation trials where the TALOS was requested to touch a table. After calibration, the average absolute positional and orientational errors of the end-effector were significantly decreased by a factor of 2.3.
    
    As  shown in previous studies of our group~\cite{bonnet2023, Villa2022}, TALOS robot suffers from relatively flexible hip-leg segments. This factor should be considered in the forthcoming studies. Moreover, to enhance the efficiency of the calibration routine, two other points could be investigated. Firstly, transitioning motions between postures can be computed offline before performing on the robot. Secondly, tactile sensors could be utilized for fast and accurate contact detection between fingers and plane in the future work.

\bibliographystyle{IEEEtran}

\bibliography{talos.bib}

% Generated by IEEEtran.bst, version: 1.14 (2015/08/26)
\begin{thebibliography}{10}
\providecommand{\url}[1]{#1}
\csname url@samestyle\endcsname
\providecommand{\newblock}{\relax}
\providecommand{\bibinfo}[2]{#2}
\providecommand{\BIBentrySTDinterwordspacing}{\spaceskip=0pt\relax}
\providecommand{\BIBentryALTinterwordstretchfactor}{4}
\providecommand{\BIBentryALTinterwordspacing}{\spaceskip=\fontdimen2\font plus
\BIBentryALTinterwordstretchfactor\fontdimen3\font minus \fontdimen4\font\relax}
\providecommand{\BIBforeignlanguage}[2]{{%
\expandafter\ifx\csname l@#1\endcsname\relax
\typeout{** WARNING: IEEEtran.bst: No hyphenation pattern has been}%
\typeout{** loaded for the language `#1'. Using the pattern for}%
\typeout{** the default language instead.}%
\else
\language=\csname l@#1\endcsname
\fi
#2}}
\providecommand{\BIBdecl}{\relax}
\BIBdecl

\bibitem{digit_robot}
E.~Ackerman, ``Humanoid robots are getting to work,'' \url{https://spectrum.ieee.org/humanoid-robots}, 2023, accessed: 30 DEC 2023.

\bibitem{Rotella14}
N.~{Rotella}, M.~{Bloesch}, L.~{Righetti}, and S.~{Schaal}, ``State estimation for a humanoid robot,'' in \emph{Proc., IEEE Int. Conf. Intell. Robots Syst.}, 2014, pp. 952--958.

\bibitem{JOUBAIR2013_compare_index}
A.~Joubair and I.~A. Bonev, ``Comparison of the efficiency of five observability indices for robot calibration,'' \emph{Mech. Mach. Theory}, vol.~70, pp. 254--265, 2013.

\bibitem{TALOS2017}
O.~Stasse, T.~Flayols, R.~Budhiraja, K.~Giraud-Esclasse, J.~Carpentier, J.~Mirabel, A.~Del~Prete, P.~Souères, N.~Mansard, F.~Lamiraux, J.-P. Laumond, L.~Marchionni, H.~Tome, and F.~Ferro, ``Talos: A new humanoid research platform targeted for industrial applications,'' in \emph{2017 IEEE-RAS Int. Conf. Humanoid Robots}, 2017, pp. 689--695.

\bibitem{bonnet2023}
V.~Bonnet, J.~Mirabel, D.~Daney, F.~Lamiraux, M.~Gautier, and O.~Stasse, ``Practical whole-body elasto-geometric calibration of a humanoid robot: Application to the talos robot,'' \emph{Robot. Auton. Syst.}, p. 104365, 2023.

\bibitem{Daney2005}
D.~Daney, Y.~Papegay, and B.~Madeline, ``Choosing measurement poses for robot calibration with the local convergencemethod and tabu search,'' \emph{Int. J. Robot. Res.}, vol.~24, no.~6, pp. 501--518, 2005.

\bibitem{Birbach2015}
O.~Birbach, U.~Frese, and B.~Bäuml, ``Rapid calibration of a multi-sensorial humanoid’s upper body: An automatic and self-contained approach,'' \emph{Int. J. Robot. Res.}, vol.~34, no. 4-5, pp. 420--436, 2015.

\bibitem{Roncone2014}
A.~Roncone, M.~Hoffmann, U.~Pattacini, and G.~Metta, ``Automatic kinematic chain calibration using artificial skin: Self-touch in the icub humanoid robot,'' in \emph{2014 IEEE Int. Conf. Robot. Autom. (ICRA)}, 2014, pp. 2305--2312.

\bibitem{Stepanova2019}
K.~Stepanova, T.~Pajdla, and M.~Hoffmann, ``Robot self-calibration using multiple kinematic chains—a simulation study on the icub humanoid robot,'' \emph{IEEE Robot. Autom. Lett.}, vol.~4, no.~2, pp. 1900--1907, 2019.

\bibitem{Stepanova2022}
K.~Stepanova, J.~Rozlivek, F.~Puciow, P.~Krsek, T.~Pajdla, and M.~Hoffmann, ``Automatic self-contained calibration of an industrial dual-arm robot with cameras using self-contact, planar constraints, and self-observation,'' \emph{Robot. Comput. Integr. Manuf.}, vol.~73, p. 102250, 2022.

\bibitem{Yamane2011}
K.~Yamane, ``Practical kinematic and dynamic calibration methods for force-controlled humanoid robots,'' in \emph{2011 11th IEEE-RAS Int. Conf. Humanoid Robots}, 2011, pp. 269--275.

\bibitem{Khusainov2017}
R.~Khusainov, A.~Klimchik, and E.~Magid, ``Humanoid robot kinematic calibration using industrial manipulator,'' in \emph{2017 Int. Conf. on Mech., Sys. and Control Eng. (ICMSC)}, 2017, pp. 184--189.

\bibitem{Tanguy2018}
A.~Tanguy, A.~Kheddar, and A.~I. Comport, ``Online eye-robot self-calibration,'' in \emph{2018 IEEE Intnationaerl Conference on Simulation, Modeling, and Programming for Autonomous Robots (SIMPAR)}, 2018, pp. 68--73.

\bibitem{khalil:hal-00362605}
W.~Khalil, P.~Lemoine, and M.~Gautier, ``{Autonomous calibration of Robots using planar points},'' in \emph{International Symposium on Robotics and Manufacturing, ISRAM'96}, vol.~3, Montpellier, France, May 1996, pp. pp. 383--388.

\bibitem{606774}
M.~Ikits and J.~Hollerbach, ``Kinematic calibration using a plane constraint,'' in \emph{Proc. - IEEE Int. Conf. Robot. Autom.}, vol.~4, 1997, pp. 3191--3196 vol.4.

\bibitem{770073}
H.~Zhuang, S.~Motaghedi, and Z.~Roth, ``Robot calibration with planar constraints,'' in \emph{Proc. 1999 IEEE Int. Conf. Robot. Autom. (Cat. No.99CH36288C)}, vol.~1, 1999, pp. 805--810.

\bibitem{Joubair2015NonkinematicCO}
A.~Joubair and I.~A. Bonev, ``Non-kinematic calibration of a six-axis serial robot using planar constraints,'' \emph{Precision Engineering}, vol.~40, pp. 325--333, 2015.

\bibitem{Detmax1974}
T.~J. Mitchell, ``An algorithm for the construction of "d-optimal" experimental designs,'' \emph{Technometrics}, vol.~16, no.~2, pp. 203--210, 1974.

\bibitem{Kamali2019}
K.~Kamali and I.~A. Bonev, ``Optimal experiment design for elasto-geometrical calibration of industrial robots,'' \emph{IEEE/ASME Trans. Mechatron.}, vol.~24, no.~6, pp. 2733--2744, 2019.

\bibitem{Sun2008}
Y.~Sun and J.~M. Hollerbach, ``Observability index selection for robot calibration,'' in \emph{2008 IEEE Int. Conf. Robot. Autom.}, 2008, pp. 831--836.

\bibitem{DHparameters}
J.~Denavit and R.~S. Hartenberg, ``{A Kinematic Notation for Lower-Pair Mechanisms Based on Matrices},'' \emph{J. Appl. Mech. Trans. ASME}, vol.~22, no.~2, pp. 215--221, 06 1955.

\bibitem{Everett1988}
L.~J. Everett and T.-W. Hsu, ``{The Theory of Kinematic Parameter Identification for Industrial Robots},'' \emph{J. Dyn. Sys., Meas., Control.}, vol. 110, no.~1, pp. 96--100, 03 1988.

\bibitem{Meggiolaro2000}
M.~Meggiolaro and S.~Dubowsky, ``An analytical method to eliminate the redundant parameters in robot calibration,'' in \emph{Proc. - 2000 IEEE Int. Conf. Robot. Autom.}, vol.~4, 2000, pp. 3609--3615 vol.4.

\bibitem{khalil_gautier_enguehard_1991}
W.~Khalil, M.~Gautier, and C.~Enguehard, ``Identifiable parameters and optimum configurations for robots calibration,'' \emph{Robotica}, vol.~9, no.~1, p. 63–70, 1991.

\bibitem{Hollerbach2008}
J.~Hollerbach, W.~Khalil, and M.~Gautier, \emph{Model Identification}.\hskip 1em plus 0.5em minus 0.4em\relax Berlin, Heidelberg: Springer Berlin Heidelberg, 2008, pp. 321--344.

\bibitem{Mirabel2016hpp}
J.~Mirabel, S.~Tonneau, P.~Fernbach, A.-K. Seppälä, M.~Campana, N.~Mansard, and F.~Lamiraux, ``Hpp: A new software for constrained motion planning,'' in \emph{2016 IEEE Int. Conf. Intell. Robots Syst. (IROS)}, 2016, pp. 383--389.

\bibitem{Nicolin2020agimus}
A.~Nicolin, J.~Mirabel, S.~Boria, O.~Stasse, and F.~Lamiraux, ``Agimus: a new framework for mapping manipulation motion plans to sequences of hierarchical task-based controllers,'' in \emph{2020 IEEE/SICE International Symposium on System Integration (SII)}, 2020, pp. 1022--1027.

\bibitem{Villa2022}
N.~A. Villa, P.~Fernbach, M.~Naveau, G.~Saurel, E.~Dantec, N.~Mansard, and O.~Stasse, ``Torque controlled locomotion of a biped robot with link flexibility,'' in \emph{2022 IEEE-RAS 21st Int. Conf. Humanoid Robots (Humanoids)}, 2022, pp. 9--16.

\end{thebibliography}
% \newpage

% \end{thebibliography}

%\appendix[Table of Identified Parameters]
% \begin{figure}[ht!]
%      \includegraphics[width=0.85\linewidth]{figures/TALOS_contact_calibration_parameters_arm.pdf}
%        \caption{Identified values of joint placement variations of the upper body by the whole-body geometric calibration process.}\label{fig:TALOS_structure}
%    \end{figure}
% \begin{figure}[ht!]
 % \centering
%      \includegraphics[width=0.85\linewidth]{figures/TALOS_contact_calibration_parameters_leg.pdf}
%        \caption{Identified values of joint placement variations of the lower body by the whole-body geometric calibration process.}\label{fig:TALOS_structure}
%    \end{figure}

%\section{Biography Section}
%If you have an EPS/PDF photo (graphicx package needed), extra braces are needed around the contents of the optional argument to biography to prevent the LaTeX parser from getting confused when it sees the complicated
 %$\backslash${\tt{includegraphics}} command within an optional argument. (You can create your own custom macro containing the $\backslash${\tt{includegraphics}} command to make things simpler here.)
 
%\vspace{11pt}

\vfill

\end{document}